\documentclass[10pt, a4paper]{article}
\usepackage{lrec}
\usepackage[utf8]{inputenc}
\usepackage{multibib}
\newcites{languageresource}{Language Resources}
\usepackage{graphicx}
\usepackage{tabularx}
\usepackage{soul}
\usepackage{float}
\usepackage{pgfplots}

\usepackage{epstopdf}

\usepackage{hyperref}
\usepackage{xstring}

\title{Monitoring Targeted Hate in Online Environments}

\name{Tim Isbister$^{1}$, Magnus Sahlgren$^{1}$, Lisa Kaati$^{1}$, Milan Obaidi$^{2}$, Nazar Akrami$^{2}$}

\address{$^{1}$Swedish Defense Research Agency (FOI), $^{2}$Uppsala University \\
         $^{1}$164 90 Stockholm, Sweden, $^{2}$Box 256, 751 05 Uppsala, Sweden\\
                     \{Tim.Isbister, Magnus.Sahlgren, Lisa.Kaati\}@foi.se, \{Milan.Obaidi, Nazar.Akrami\}@psyk.uu.se\\}

\abstract{Hateful comments, swearwords and sometimes even death threats are becoming a reality for many people today in online environments. This is especially true for journalists, politicians, artists, and other public figures. This paper describes how hate directed towards individuals can be measured in online environments using a simple dictionary-based approach. We present a case study on Swedish politicians, and use examples from this study to discuss shortcomings of the proposed dictionary-based approach. We also outline possibilities for potential refinements of the proposed approach.}

\begin{document}
\maketitleabstract
\vspace{.3\baselineskip}

\section{Introduction}
Digital environments provide an enormously large and accessible platform for people to express a broad range of behavior --- perhaps even broader than what can be expressed in real world environments, due to the lack of social accountability in many digital environments. Hate and prejudice are examples of such behaviors that are overrepresented in digital environments. Hate messages in particular are quite common, and have increased significantly in recent years. 
In fact, many, if not most, digital newspapers have closed down the possibility to comment on articles since the commentary fields have been overflowing with hate messages and racist comments \cite{gardiner}. To many journalists, politicians, artists, and other public figures, hate messages and threats have become a part of daily life. A recent study on Swedish journalists showed that almost 3 out of 4 journalists received threats and insulting comments through emails and social media~\cite{nilsson}.

Several attempts to automatically detect hate messages in online environments have been made. For example, Warner and Hirschberg \shortcite{Warner:2012} use machine learning coupled with template-based features to detect hate speech in user-generated web content with promising results. Wester et al.~\shortcite{WesOvrVel16} examine the effects of various types of linguistic features for detecting threats of violence in a corpus of YouTube comments, and find promising results even using simple bag-of-words representations. On the other hand, Ross et al.~\shortcite{ross} examine the reliability of annotations of hate speech, and find that the annotator agreement is very low, indicating that hate speech detection is a very challenging problem. The authors suggest that hate speech should be seen as a continuous rather then as a binary problem, and that detailed instructions for the annotators are needed to improve the reliability of hate speech annotation. Waseem and Hovy \shortcite{Waseem:2016} examine the effect of various types of features on hate speech detection, and find that character n-grams and gender information provide the best results. Davidson et al.~\shortcite{DavidsonWMW17} argues that lexical methods suffer from low precision and aims to separate hate speech from other instances of offensive language. Their results show that while racist and homophobic content are classified as hate speech, this is not the case for sexist content, which illustrates the challenge in separating hate speech from other instances of offensive language. 

The apparent lack of consensus regarding the difficulty of the hate speech detection problem suggests that the problem of hate speech detection deserves further study. This paper contributes to the discussion in two ways. Firstly, we provide a psychological perspective on the concept of hate. Secondly, we present a study of the advantages and disadvantages of using the arguably simplest possible approach to hate speech detection: that of counting occurrences of keywords based on dictionaries of terms related to hate speech. The main goal of this paper is to provide a critical discussion about the possibility of monitoring targeted hate in online environments.

This paper is outlined as follows. Section~2 discusses the psychological aspects of hate and how hate messages can have various level of severity. Section~3 presents a dictionary-based approach to measure hate directed towards individuals. Section~4 provides a case study where we analyze hate speech targeted towards 23 Swedish politicians on immigration-critical websites, and discuss challenges and directions for future work. Finally, Section~5 provides some concluding remarks.

\section{On hate \label{sec:hate}}
In the psychological literature hate is thought to be a combination to two components: one cognitive and one emotional \cite{Sternberg}. The cognitive component can be threat perceptions caused for example by out-group members, but it can also involve devaluation or a negative view of others. The emotional component on the other hand involves emotions such as contempt, disgust, fear, and anger that are generally evoked by the cognitive component. Defined in this way, hates shares much with prejudice, which is defined as negative evaluations or devaluations of others based on their group membership. Like hate, prejudice is argued to be consisting of a cognitive component (stereotypes about others), an emotional component (dislike of others) and a behavioral component (acting in accordance with the emotional and cognitive component \cite{Allport}). Hate, like prejudice, functions as the motivational force when people behave in harmful ways toward others. 

\begin{table*}[h]
\begin{center}
\begin{tabular}{lllc}
{\bf Category} & {\bf Sample terms (ENG)} & {\bf Sample terms (SWE)} & {\bf Normalized frequency per category}\\
\hline
Swearword&fuck, shit, gay & fan, skit, bög & $0.00137$\\
Anger& is crazy, idiot, enemy & är galen, idiot, fiende & $0.00106$\\ 
Naughtiness& clown, is an idiot, stupid & clown, är en idiot, knäpp & $0.00076$\\
General threat& kidnap, be followed, hunt & kidnappa, bör förföljas, jaga & $0.00068$\\
Death threat& should be killed, ruin, bomb & borde dödas, utrota, bomba & $0.00031$\\
Sexism& whore, bitch, should be raped& hora, subban, borde våldtas & $0.00005$\\
\end{tabular}
\end{center}
\caption{Different categories of hate with representative terms and normalized frequency.}
\label{tab:cat}
\end{table*}

Hate is commonly directed toward individuals and groups but it is also expressed toward other targets in the social world. For example, it is common that hate is expressed toward concepts (e.g.~communism) or countries (e.g.~USA). It is important to note however that there is some disagreement about not only the definition but also the behavioral outcomes of hate. For example, while some see hate leading to behavioral tendencies such as withdrawal caused by disgust or fear, others see hate as the manifestation of anger or rage, which lead one to approach, or attack, the object of hate \cite{Rozman}.

Dealing with digital environments, the disagreement about behavioral tendencies might seem less relevant. Specifically, withdrawal caused by disgust or fear in the real world is not the same in digital environment where withdrawal would not be necessary --- or approach would not be a direct threat to wellbeing. Acknowledging the disagreements noted above, we aim to examine hate messages with various level of severity varying between swearwords directed to individuals to outright death threats. 

\section{Monitoring hate}
\label{sec:monitoring}
This work focuses on detecting hate messages and expressions directed towards individuals. The messages can have various level of severity with respect to individual integrity and individual differences in perception of threat. More specifically, we examine six different categories: anger, naughtiness, swearwords, general threats, and death threats. While the two categories naughtiness and anger may overlap in some aspects, they were aimed to capture different expressions and causes of hate speech, with naughtiness indicating to the speaker's tendency to misbehave and generally express naughtiness toward others, and anger being an emotional state triggered by something in the surrounding and leading to the expression of anger (and/or naughtiness) towards a person. We also include sexism (degradation of women), since it is commonly used for devaluative purposes. Each category is represented by a dictionary of terms, as exemplified in Table \ref{tab:cat}. Our study focuses on Swedish data, but to ease understanding we have translated some of the words to English. Note that the dictionaries may contain both unigrams and multiword expressions.

The dictionaries are constructed in a manner similar to Tulkens et al.~\shortcite{Tulkens:tacos,Tulkens:2016}; human experts (psychologist and computer scientist) manually study a large number of posts from the text domain of interest (see further Section \ref{sec:data}) and record significant words and phrases. In order to improve the recall of the dictionaries, a word embedding is then used to suggest other relevant terms to the experts. This is done by simply computing the 15 nearest neighbors in the embedding space to each term in the dictionaries. For each term suggestion, the expert has the choice to either include or reject the term suggestion. We note that it is also possible to cast the term suggestion task as an active learning problem, in which a classifier is iteratively refined to identify useful term suggestions based on the expert's feedback \cite{Gyllensten:2018}.

As embedding, we use Gensim's \cite{rehurek_lrec} implementation of the Continuous Bag of Words (CBOW) model \cite{Mikolov:2013}, which builds word vectors by training a 2-layer neural network to predict a target word based on a set of context words. The network learns two sets of vectors, one for the target terms (the embedding vectors), and one for context terms. The objective of the network is to learn vectors such that their dot product correspond to the log likelihood of observing word pairs in the training data. We use default parameters for the embeddings, with a window size set to 5. The embeddings are trained on a collection of immigration-critical websites, further discussed in Section \ref{sec:data}. Note that the embedding method does not handle multiword units in any special way; if multiword units are to be included in the analysis, they need to be incorporated in the data as a preprocessing step.

The expanded dictionaries are used to detect and monitor hate by simple frequency counting; if a term from one of the dictionaries occurs in the vicinity of a mention of a target individual, we increment the count for that category. This is arguably the simplest possible approach to hate speech monitoring, and many types of refinements are possible, such as weighting of the dictionary entries \cite{Eisenstein17}, handling of negation \cite{reitan-EtAl:2015:WASSA}, and scope detection. We will return to a more detailed discussion of problems with the proposed approach in Section \ref{sec:error}. At this point, we note that one possible advantage of using such a simple approach is its transparency; it is easy to understand a simple frequency counter for a non-technical end user.

Of course, transparency and comprehensibility are useless if the method generates an excessive amount of false positives. The only way for us to control the precision of the frequency counting is to delimit the context within which occurrences of dictionary terms are counted; a narrow context window spanning something like one to three words around a target individual's name will reduce the probability that a term from one of the dictionaries refers to something other than the target name. In the following case study, we opt for the most conservative approach and use a context of only one term on each side of the target name. 
 
\section{Case study}
\label{sec:case}

To exemplify the dictionary-based approach, we have examined the expression of the different categories of hate toward 23 national-level politicians (10 males and 13 females). Studying national-level politicians in Sweden is timely as we are approaching the Swedish parliament election in September 2018. There have also been recent alarms on politicians threatening to leave politics because of an increasing amount of hate being expressed in recent years. Our analyses are based on text from commentary fields on immigration critical websites from September 2014 to December 2017. The time period was chosen to cover a single electoral period in the Swedish parliament. 

As target names, we use the full names of the politicians. This is obviously a crude simplification that severely affects the recall of the approach, since people are often referred to using only their first name, a pronoun, or, in the data we studied, some negative nickname or slur. As an example, the Swedish prime minister, Stefan Löfven, is often referred to in online discussions as ``svetsarn'' (the welder), or using negative nicknames such as ``Röfven'', which is a paraphrase of ``röven'' (in English ``the ass'').

\subsection{Data}
\label{sec:data}
In Sweden, as well as in several other European countries, there has been a recent surge in activity and formation of movements that are critical of immigration. These immigration-critical groups show a high interactivity on social media and on websites. In Sweden, there are several digital immigration-critical milieus with a similar structure: articles published by editorial staff and user-generated comments. The commentary fields are not moderated, which makes the comments an important scene to express hate toward journalists, politicians, artists, and other public figures. The comment section allows readers to respond to an editorial article instantly. The editorial articles generally focuses on topics such as crimes, migration, politics and societal issues. The websites that we have studied are listed in Table~\ref{table:sites}. For each website, we have downloaded all comments between 2014/09/01 to 2017/10/01. Note that the embeddings used for term suggestions are also trained on this data.

\begin{table}
\begin{center}
\begin{tabular}{l r r}
\bf{Website}& \bf{\# comments} & {\bf \# words}\\
\hline
\texttt{avpixlat.info} & 2 904 933 & 99 472 281 \\
\texttt{nordfront.se} & 89 495 & 3 125 218 \\
\texttt{nyatider.nu} & 2 176 & 124 949 \\
\texttt{motgift.nu} & 1 380 & 68 992 \\
\texttt{nordiskungdom.com} & 117 & 6 530 \\
\end{tabular}
\caption{The websites included in our study.}
\label{table:sites}
\end{center}
\end{table}

\begin{table}[h]
\begin{center}
\begin{tabular}{l r }
\bf{Name}& \bf{Mentions}\\
\hline
Stefan Löfven & 10 663\\
Morgan Johansson& 3 142\\
Margot Wallström& 2 681\\
Magdalena Andersson& 1 931\\
Ylva Johansson &1 524\\
Gustav Fridolin &1 113\\
Alice Bah Kuhnke &567\\
Peter Eriksson & 248\\
Peter Hultqvist & 228\\
Isabella Lövin& 184\\
Mikael Damberg & 169\\
Ardalan Shekarabi & 158\\
Åsa Regnér &136\\
Ann Linde & 128\\
Annika Strandhäll& 98\\
Ibrahim Baylan & 61\\
Per Bolund & 48\\
Anna Ekström & 36\\
Heléne Fritzon& 36\\
Helene Hellmark Knutsson & 14\\
Karolina Skog & 11\\
Sven-Erik Bucht &8\\
\end{tabular}
\caption{Number of times each Swedish minister is mentioned in the comments during the time period.}
\label{table:names}
\end{center}
\end{table}

\subsection{Results}

Table \ref{table:names} shows the how many times each minister is mentioned in the comments with his or hers full name during the given time period. Obviously, the Prime Minister Stefan Löfven is the most frequently mentioned politician, with more than 10,000 mentions during the analyzed period. The second most mentioned politician in the studies data is Morgan Johansson, the Swedish Minister of Justice and Home Affairs, and the third most mentioned minister is Margot Wallström, Minister for Foreign Affairs.

\begin{figure*}
\label{fig1}
	\includegraphics[scale=0.41]{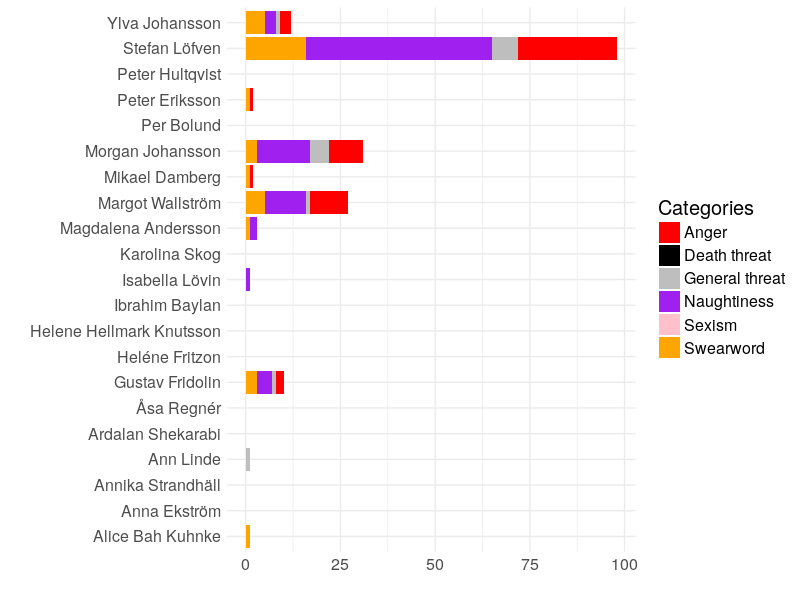} \includegraphics[scale=0.32]{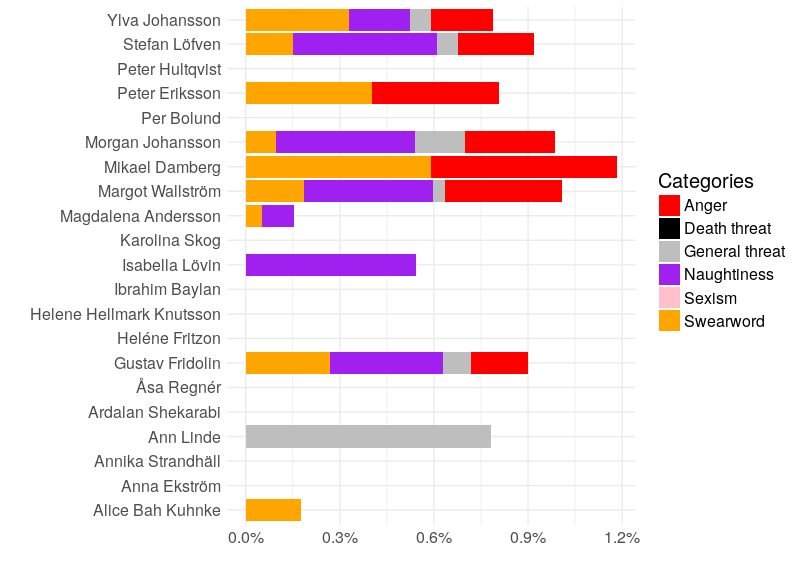} 
\caption{Amount of hate contexts for Swedish ministers (using only the preceding and succeeding terms). The left figure shows simple frequency counts of hate terms, while the right figure shows proportions (i.e.~counts divided by the total number of mentions).}
\label{fig:res}
\end{figure*}

Figure \ref{fig:res} (next page) shows the amount of hate towards the Swedish ministers. The left figure shows simple frequency counts of hate terms in the immediate vicinity of each target name, while the right figure shows the proportions of targeted hate toward the Swedish ministers, calculated as the frequency of each hate category in the context of each politician, divided by the total number of mentions for that politician. In both figures, it is obvious that naughtiness (in purple) is the most frequent category for the politicians as a group, followed by anger (in red), swearwords (in yellow) and general threat (in gray). We do not see any sexism and no explicit death threats in our data, most likely due to the very narrow context used in these experiments. 

Figure \ref{fig:res} shows that the most frequently mentioned ministers are also those who receive the most hate in the data we have studied. However, when looking at the proportions of hateful comments for each minister, we see that the most mentioned politician (Stefan Löfven) is not the minister with the proportionally most hateful comments. This is instead Mikael Damberg, the Minister for Enterprise and Innovation. However, Damberg is only mentioned 169 times in the data, and a mere 1.18\% of these contain hate; that is, only 2 mentions of 169. It is a similar situation with Ann Linde, the Minister for EU Affairs and Trade, who has the proportionally most general threats in her mentions, but this is based on only 1 mention out of 128. Isabella Lövin, the Minister for International Development Cooperation, is the target of the proportionally most naughtiness, but also in this case, this is only 1 mention out of 184.

\subsection{Discussion}\label{sec:error}
The results in Figure \ref{fig:res} demonstrate that even with such a simple and na\"ive method as the one used in this paper, it is possible to do a general and rudimentary form of threat assessment based on mentions in social media data. The method is sufficiently simple to be adaptable to many different scenarios, and sufficiently transparent for end-users to understand. However, we do pay a price for the simplicity.

As we noted in the last section, expressions of hate seem to correlate with frequency of mention (at least in the data we have studied). This makes the left part of Figure \ref{fig:res} less interesting. On the other hand, counting proportions, as we do in the right part of the figure, risks overestimating the significance of very rare events. A perhaps more useful measure might be to calculate deviations from the expected amount of hateful comments for each minister. As an example, Morgan Johansson is mentioned 3~142 by his full name in our data. Based on the normalized category frequencies in Table \ref{table:names}, we should expect that 4 of these mentions contain swearwords, 3 contain anger, 2 contain naughtiness, and 2 contain general threat. Looking at the actual frequency counts, we see that 3 mentions contain swearwords, 8 contain anger, 14 contain naughtiness, and 5 contain general threat. For the last three categories, the actual counts are much higher than would be expected, indicating that these are significant measurements. 

Table \ref{tab:dev} (next page) shows the deviations from expected counts per category for each minister. The deviation is computed as the actual counts minus the expected counts:
\begin{equation}
\#(m,c)-\bigg(\frac{\#(c)}{T}\cdot\#(m)\bigg)
\end{equation}
\noindent
where $\#(m,c)$ is the actual co-occurrence count of a minister and a category, $\frac{\#(c)}{T}$ is the relative frequency of a category in the data $\#(c)$ is the frequency of the category and $T$ is the total number of words in the data), and $\#(m)$ is the frequency of mention of a minister. 

This is a obviously a severely oversimplified probabilistic model, but it does provide useful information. We note that the columns for death threats and sexism only contain negative or zero values, which indicates that no significant death threats or sexism is being expressed towards the ministers in the data. Two ministers have higher general threats than can be expected, and a few more have higher swearwords and anger, but the deviations for these categories in our data are not very large. The highest deviation in our study is the naughtiness category for the prime minister, which indicates that he is the subject of a significant amount of negative remarks in the data we have studied. Another potentially interesting observandum is the combination of categories that have positive deviations for the different ministers. To take two examples, Morgan Johansson has positive deviations for anger, naughtiness and general threat, while Ylva Johansson has positive deviations for swearwords, anger and naughtiness.  One might hypothesize that the combination of anger and general threat deserves more attention than the combination of swearwords and naughtiness.

\begin{table*}[h!]
\centering
\label{actual_minus_expected}
\begin{tabular}{lcccccc}
 {\bf Person} & {\bf Swearword} & {\bf Anger} & {\bf Naughtiness} & {\bf General threat} & {\bf Death threat} & {\bf Sexism}  \\
 \hline
 Stefan Löfven & $0.98$ & $3.29$ & ${\bf 16.49}$ & $-2.65$ & $-3.15$ & $-0.46$ \\
 Morgan Johansson & $-1.16$ & $2.82$ & $2.77$ & $2.32$ & $-0.93$  & $-0.14$ \\
 Margot Wallström & $1.5$ & $2.32$ & $3.12$ & $-1.41$ & $-0.79$ & $-0.12$ \\
 Magdalena Andersson & $-1.56$ & $-1.96$ & $0.63$ & $-1.03$ & $-0.57$ & $-0.08$ \\
 Ylva Johansson & $2.95$ & $1.43$ & $1.9$ & $-0.83$ & $-0.46$ & $-0.07$ \\
 Gustav Fridolin & $1.51$ & $-0.14$ & $2.2$ & $-0.6$ & $-0.33$ & $-0.05$ \\
 Alice Bah Kuhnke & $0.24$ & $-0.58$ & $-0.4$ & $-0.3$ & $-0.17$ & $-0.02$ \\
 Peter Eriksson & $0.67$ & $0.74$ & $-0.18$ & $-0.13$ & $-0.08$ & $-0.01$ \\
 Peter Hultqvist & $-0.29$ & $-0.22$ & $-0.15$ & $-0.12$ & $-0.06$ & $-0.01$ \\
 Isabella Lövin & $-0.24$ & $-0.18$ & $0.87$ & $-0.1$ & $-0.05$ & $-0.01$ \\
 Mikael Damberg & $0.77$ & $0.83$ & $-0.12$ & $-0.09$ & $-0.05$ & $-0.01$ \\
 Ardalan Shekarabi & $-0.21$ & $-0.16$ & $-0.11$ & $-0.08$ & $-0.05$ & $-0.01$ \\
 Åsa Regnér & $-0.18$ & $-0.14$ & $-0.1$ & $-0.07$ & $-0.04$ & $-0.01$ \\
 Ann Linde & $-0.17$ & $-0.13$ & $-0.09$ & $0.93$ & $-0.04$ & $-0.01$ \\
 Annika Strandhäll & $-0.13$ & $-0.1$ & $-0.07$ & $-0.05$ & $-0.03$ & $0$ \\
 Ibrahim Baylan & $-0.08$ & $-0.06$ & $-0.04$ & $-0.03$ & $-0.02$ & $0$ \\
 Per Bolund & $-0.06$ & $-0.05$ & $-0.03$ & $-0.02$ & $-0.01$ & $0$ \\
 Anna Ekström & $-0.05$ & $-0.04$ & $-0.03$ & $-0.02$ & $-0.01$ & $0$ \\
 Heléne Fritzon & $-0.01$ & $-0.01$ & $-0.01$ & $-0.01$ & $0$ & $0$ \\
 Helene Hellmark Knutsson & $-0.02$ & $-0.01$ & $-0.01$ & $-0.01$ & $0$ & $0$ \\
 Karolina Skog & $-0.01$ & $-0.01$ & $-0.01$ & $-0.01$ & $0$ & $0$ \\
\end{tabular}
\caption{Deviation from expected counts per category for each minister. Positive scores indicate that the actual count is higher than the expected count.}
\label{tab:dev}
\end{table*}



The perhaps most obvious drawback of the approach used in this paper is that it will only detect hate in direct relation to a full name, but not in relation to pronouns or slang expressions referring to the person in question; i.e.~the approach suffers from a lack of coreference resolution. This will obviously affect the recall of the method, which is a serious shortcoming that risks missing critical mentions. In the present analysis, we have no idea whether the lack of death threats in our results is due to an actual absence of death threats in the data, or whether it is due to omissions in the analysis. 

Although we delimit the context as much as possible to only include the preceding and succeeding terms, our results are still affected by false positives. There are three basic error types for false positives in our analysis. One is negated statements, such as (hate term in boldface):

\vspace{20pt}
\texttt{jag tror inte Stefan Lövfen är dum}\\
(I don't think Stefan Lövfen {\bf is stupid})
\vspace{10pt}

Handling negations is a well-known issue in both information retrieval and sentiment analysis, and one could think of several different ways to deal with negations. The perhaps most simple method is to use a {\em skip} or {\em flip} function that skips a sequence of text when having encountered a negation, or simply flips the sentiment of the negated text \cite{Choi:2009}. It is of course also necessary to determine the scope of the negation, which is a non-trivial problem in itself \cite{Lazib}.

Another error type in our analysis is quotes, such as:

\vspace{10pt}
\texttt{vi har varit naiva [sa] Stefan Löfven}\\
(we have been {\bf naive} [said] Stefan Löfven)
\vspace{10pt}

The ``said'' is implicit, and is signaled by quotation marks and punctuation in the original data. However, when using aggressive tokenization, such punctuation is normally removed, which leads to the above type of errors. Retaining punctuation would obviously be one way to prevent such errors. Another possibility is to use a dependency parse of the data, which would rearrange the context according to the dependency structure. ``Naive'' would then be closer to ``we'' than to ``Stefan Löfven''. 

A third error type that is related to the previous one is misinterpreting (or ignoring) the semantic roles of the proposition. Consider the following examples:

\vspace{10pt}
\texttt{låt regeringen med Stefan Lövfen hota med nyval}\\
(let the government with Stefan Lövfen {\bf threaten} with new election)
\vspace{10pt}

\vspace{10pt}
\texttt{vi skiter i om du blir förbannad Stefan Lövfen}\\
(we don't care if you get {\bf upset} Stefan Lövfen)
\vspace{10pt}

Stefan Löfven is not the target of hate in neither of these cases. Instead, he (or in the first case, he and the Swedish government) is the {\em agent} of the predicates ``threatened'' vs.~``upset''. In order to resolve agency of the predicates, we would need to do semantic role labeling, which assigns a semantic role to each participant of a proposition. Identifying the agent of the predicate becomes even more important when increasing the context size, since it will also increase the number of false positives when only counting occurrences of hate terms.

\section{Conclusion}
\label{sec:concl}
In this paper, we have aimed to measure how online hate is directed toward national-level politicians in Sweden. This is an important and timely endeavor because the expression of online hate is becoming increasingly pervasive in online forums, especially toward this specific group. The expression of hate has shown to have downstream consequences not only for individuals who are targeted, but also for our democratic society and core liberal values. Recent studies show that the frequent exposure to hate speeches does not only lead to increased devaluation and prejudice \cite{Soral:2017}, but may also increase dehumanization of the targeted group \cite{Fasoli:2016}. Dehumanization in return makes the targeted groups or individuals seem less than human, legitimizing and increasing the likelihood of violence \cite{Rai:2017}. Moreover, online hate does not only play a significant role in shaping people's attitudes and beliefs toward certain groups, but it also have far-reaching consequences for societies in general, such as increasing tendency to violating social norms and threatening democratic core values. 

As we mentioned in the introduction, many digital newspapers in Sweden and other countries have closed down the possibility to comment on articles due to the degree of hate expressed by some users. This is a clear example of how online hate restricts and threatens one of the core values of democracy. That is the freedom to express your views and opinions. To prevent such harmful effects it is important to monitor and measure how and toward whom hate is expressed online.

The second aim of this study was to address some of the gaps in the field. As noted in the introduction, the contemporary approaches to measuring online hate are marked by the apparent lack of consensus regarding the difficulty of the hate speech detection. The approach for monitoring targeted hate that we have described in this work is a simple yet powerful way to understand hate messages directed toward individuals. The strength of this method lies in its simplicity and transparency, and perhaps also for having more conservative criteria that reduces the number of false positives. We have also identified a number of ways to improve the method, including the use of {\bf coreference resolution}, handling of {\bf negation}, context refinement using {\bf dependency parsing}, and agency detection using {\bf semantic role labeling}.

The trade-off between complexity and performance, and between recall and precision, are challenging dilemmas for law enforcement and other end users of hate monitoring tools. Acknowledging these dilemmas, future improvements of hate monitoring should be directed toward the optimal cut-off where usefulness for law enforcement can meet ease of conduct when it comes to analyzing data. 

\section{References}
\bibliographystyle{lrec}
\bibliography{xample}
\end{document}